\documentclass[10pt,final,journal,twocolumn]{IEEEtran}
\usepackage{amsmath,amssymb,color,amsmath, graphicx, float, subcaption, mathrsfs,color}
\usepackage[ruled]{algorithm2e} 
\usepackage{bm}
\usepackage{epstopdf}
\usepackage[font={small}]{caption}
\usepackage[font={footnotesize}]{subcaption}
\usepackage{cite}
\usepackage{amsthm}
\usepackage{url}

\newtheorem{thm}{Theorem}
\newtheorem{defn}{Definition}

\ifCLASSINFOpdf
\else
\fi
\begin{document}
\title{Lautum Regularization for \\ Semi-supervised Transfer Learning}

\author{Daniel Jakubovitz,
        Miguel R. D. Rodrigues,
        and~Raja Giryes
\thanks{Daniel Jakubovitz and Raja Giryes are with the School of Electrical Engineering, Tel Aviv University, Tel Aviv 69978, Israel. (email: danielshaij@mail.tau.ac.il,
raja@tauex.tau.ac.il)}
\thanks{Miguel R. D. Rodrigues is with the Department of Electronic and Electrical Engineering, University College London, London, United Kingdom. (email: m.rodrigues@ucl.ac.uk)}}


\maketitle

\begin{abstract}
Transfer learning is a very important tool in deep learning as it allows propagating information from one "source dataset" to another "target dataset", especially in the case of a small number of training examples in the latter. Yet, discrepancies between the underlying distributions of the source and target data are commonplace and are known to have a substantial impact on algorithm performance.
In this work we suggest a novel information theoretic approach for the analysis of the performance of deep neural networks in the context of transfer learning.
We focus on the task of semi-supervised transfer learning, in which unlabeled samples from the target dataset are available during the network training on the source dataset. Our theory suggests that one may improve the transferability of a deep neural network by imposing a Lautum information based regularization that relates the network weights to the target data. We demonstrate the effectiveness of the proposed approach in various transfer learning experiments. 
\end{abstract}


%
\IEEEpeerreviewmaketitle

\section{Introduction}
\IEEEPARstart{M}{achine} learning algorithms have lately come to the forefront of technological advancements, providing state-of-the-art results in a variety of fields \cite{Goodfellow16Deep}.
However, alongside their incredible performance, these methods suffer from sensitivity to data discrepancies - any inherent difference between the training data and the test data may result in a substantial decrease in performance.
Moreover, to obtain good performance a large amount of labeled data is necessary for their training. Such a substantial amount of labeled data is often either very expensive or simply unobtainable.

\begin{figure*}
    \centering
    \begin{subfigure}[c]{0.8\textwidth}
    \centering
    \includegraphics[width=\columnwidth]{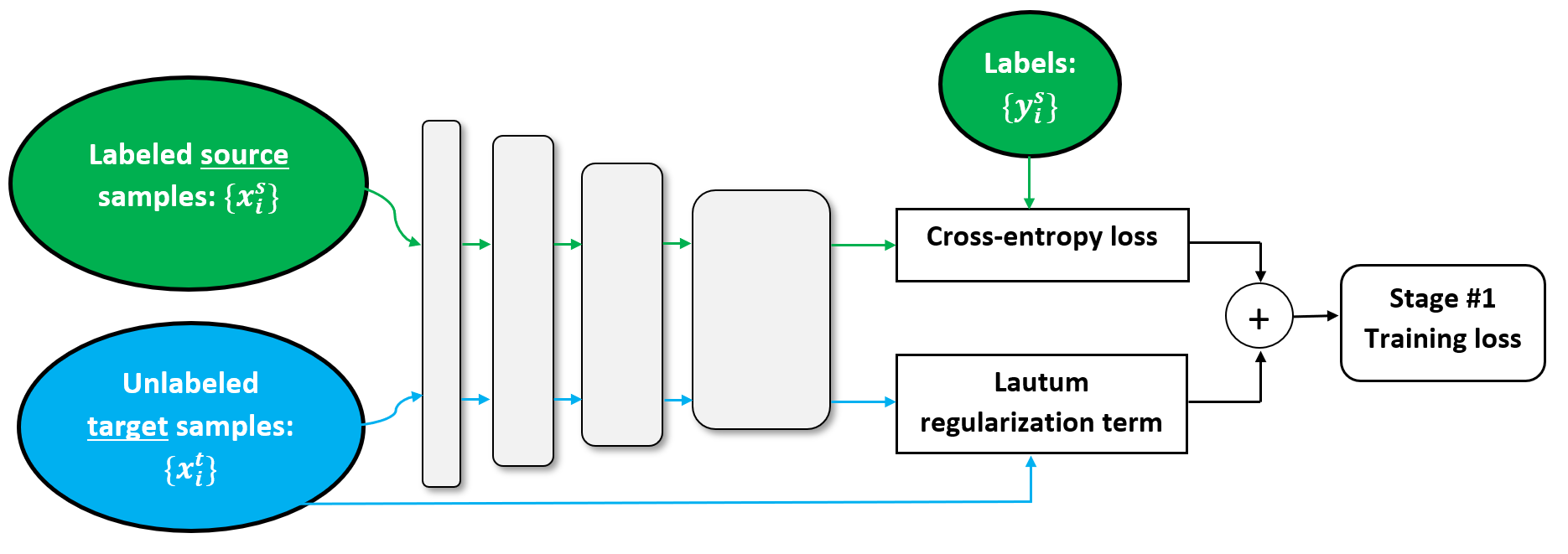}
    \caption{Pre-transfer training stage.}
    \end{subfigure}
    \hfill
    \begin{subfigure}[c]{0.8\textwidth}
    \centering
    \includegraphics[width=\columnwidth]{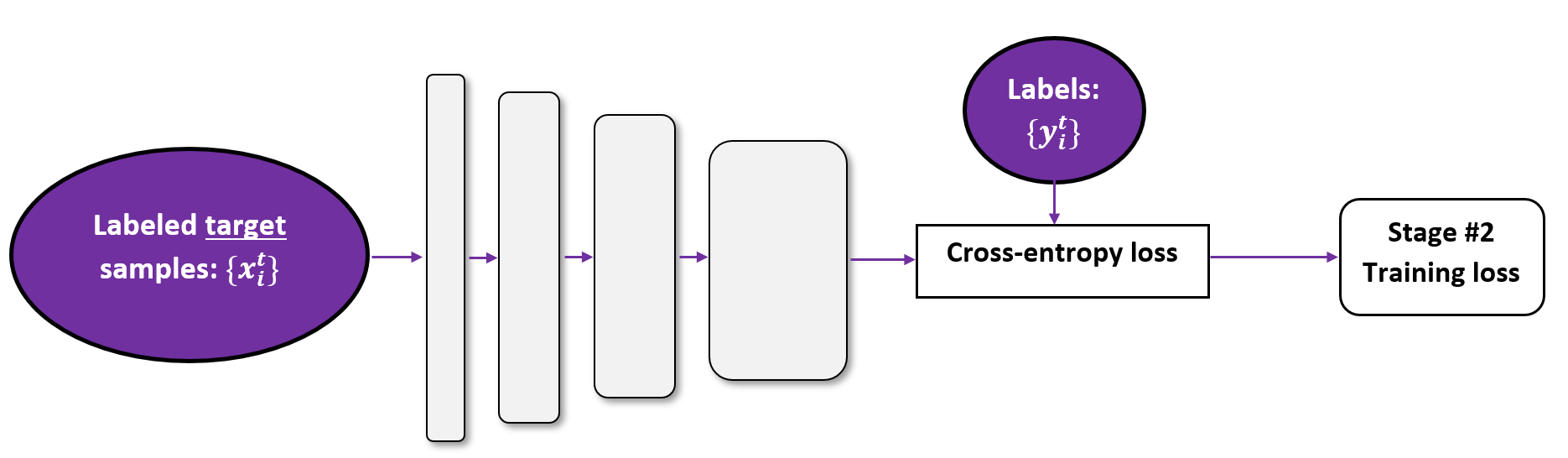}
    \caption{Post-transfer training stage.}
    \end{subfigure}
    \caption{Our semi-supervised transfer learning technique applying Lautum regularization. Omitting the blue part in the first training stage (top) gives standard transfer learning.}
    \label{fig:Our_training_technique}
\end{figure*}

One popular approach to mitigate this issue is using "transfer learning", where training on a small labeled "target" dataset is improved by using information from another large labeled "source" dataset of a different problem.
A common method for transfer learning uses the result of training on the source as initialization for training on the target, thereby improving the performance on the latter \cite{Donahue_2014_Decaf}.

Transfer learning has been the focus of much research attention along the years. Plenty of different approaches have been proposed to encourage a more effective transfer from a source dataset to a target dataset, many of them aim at obtaining better system robustness to environment changes, so as to allow an algorithm to perform well even under some variations in the settings (e.g. changes in lighting conditions in computer vision tasks). Sometimes this is achieved at the expense of diminishing the performance on the original task or data distribution.
Other works take a more targeted approach and directly try to reduce algorithms' generalization error by decreasing the difference in their performance on specific source and target datasets \cite{Xuhong_2018_Explicit_TL}.
In addition, it is often the case that the target dataset has a large number of samples, though only a few of those samples are labeled. In this scenario a semi-supervised learning approach could prove to be beneficial by making good use of the available unlabeled samples for training.


In this work we focus on the task of semi-supervised transfer learning. The problem we address is related to the field of domain adaptation, however we make a distinction between domain adaptation and transfer learning, where the former refers to the case of two sources of data with the same content (e.g. the MNIST $\rightarrow$ SVHN case) whereas the latter refers to the case of two sources of data which are completely different in both content and "styling". Another relevant difference is that labeled data from the target distribution is typically available in the transfer learning case, yet less so in the domain adaptation case. 

Plenty of works exist in the literature on transfer learning, semi-supervised learning and using information theory for the analysis of machine learning algorithms.
The closest work to ours is \cite{AchilleS18_InvarianceDisentanglement} in which an information theoretic approach is used in order to decompose the cross-entropy \emph{train} loss of a machine learning algorithm into several separate terms. However, unlike this work we propose a different decomposition of the cross-entropy \emph{test} loss and make the relation to semi-supervised transfer learning. \\

{\bf Contribution.}
We consider the case of semi-supervised transfer learning in which plenty of labeled examples from a source distribution are available along with just a few labeled examples from a target distribution; yet, we are provided also with a large number of unlabeled samples from the latter.
This setup combines transfer learning and semi-supervised learning, where both aim at obtaining improved performance on a target dataset with a small number of labeled examples.
In this work we suggest to combine both methodologies to gain the advantage of both of them.
This setting represents the case where the learned information from a large labeled source dataset is used to obtain good performance when transferring to a mostly unlabeled target set, where the unlabeled examples of the target are available at the training time on the source.

To do so, we provide a theoretical derivation that leads to a novel semi-supervised technique for transfer learning.
We take an information theoretic approach to examine the cross-entropy test loss of machine learning methods. We decompose the loss to several different terms that account for different aspects of its behavior.
This derivation leads to a new regularization term, which we call "Lautum regularization" as it relies on the maximization of the Lautum information \cite{Verdu_08_Lautum} between unlabeled data samples drawn from the target distribution and the learned model weights.
Figure~\ref{fig:Our_training_technique} provides a general illustration of our approach.

We corroborate the effectiveness of our approach with experiments of semi-supervised transfer learning for neural networks on image classification tasks.
We examine the transfer in two cases: from the MNIST dataset to the notMNIST dataset (which consists of the letters A-J in grayscale images) and from the CIFAR-10 dataset to 10 specific classes of the CIFAR-100 dataset.
We compare our results to three other methods: (1) Temporal Ensembling (TE) \cite{Laine_18_TemporalEnsembling}, a state-of-the-art method for semi-supervised training which we apply in a transfer learning setup. We examine TE both by itself and combined with Lautum regularization; (2) the Multi-kernel Maximum Mean Discrepancy (Mk-MMD) method \cite{Mk_MMD}, which is popular in semi-supervised transfer learning; (3) standard transfer learning which does not use any of the unlabeled samples.
The advantage of our method is demonstrated in our experimental results as it outperforms the other compared methods.

\section{Related Work}
\label{sec:Related Work}
Plenty of works exist in the literature on transfer learning, semi-supervised learning and using information theory for the analysis of machine learning algorithms. We hereby overview the ones most relevant to our work. \\

\textbf{Transfer learning.}
Transfer learning \cite{JialinPan_10_TransferSurvey, Tan_2018_DL_Transfer_Survey} is a useful training technique when the goal is to adapt a learning algorithm, which was trained on a source dataset, to perform well on a target dataset that is potentially very different in content compared to the source. This technique can provide a significant advantage when the number of training samples in the source dataset is large compared to a small number in the target, where the knowledge extracted from the source dataset may be relevant also to the target.

The work in \cite{Jason_2014_HowTransferable} relates to a core question in transfer learning: which layers in the network are general and which are more task specific, and precisely how transferability is affected by the distance between two tasks.
In a recent work \cite{Lampinen_19_TransferLinear} an analytic theory of how knowledge is transferred from one task to another in deep linear networks is presented. A metric is given to quantify the amount of knowledge transferred between a pair of tasks.

Practical approaches for improving performance in transfer learning settings have been proposed in many works.
In \cite{Xuezhi_2014_Flexible_Transfer}, transfer learning in the context of regression problems is examined. A transfer learning algorithm which does not assume that the support of the target distribution is contained in the support of the source distribution is proposed. This notion leads to a more flexible transfer.
In \cite{Wei_18_Learning_to_transfer} the framework of "learning to transfer" is proposed in order to leverage previous transfer learning experiences for better transfer between a new pair of source and target datasets.
In \cite{Zamir_2018_Taskonomy} the structural relation between different visual tasks is examined in the feature space. The result is a taxonomic map that enables a more efficient transfer learning with a reduced amount of labeled data.

All of the above works differ from ours in their core approach - they address the discrepancy between the source and target data either in the input space or in the feature space, yet disregard the effect of the chosen loss function and its impact on the mitigation of this discrepancy.
In contrast, our work is focused on the mathematical analysis of the cross-entropy loss which is commonly used in classification tasks. \\

\textbf{Semi-supervised learning.}
Semi-supervised learning \cite{Zhu_06_SemiSupervisedSurvey} is typically used when there is little labeled data for training, yet more unlabeled data is available.
The literature on semi-supervised training is vast and describes a variety of techniques for performing effective semi-supervised learning that would make good use of the available unlabeled data in order to improve the model performance.
Most of these techniques rely on projecting the relation between the available labeled samples and their labels to the unlabeled samples and the model's predicted labels for them.

In \cite{Yves_05_SemiSupervised_Entropy} minimum entropy regularization is proposed. This technique modifies the cross-entropy loss used for training in order to encourage a deep neural network to make confident predictions on unlabeled data.
In \cite{HausserMC_17_LearningByAssociation} a new framework for semi-supervised training of neural networks called "associative learning" is proposed. In this framework "associations" are made between the embeddings of the available labeled data and the unlabeled data. An optimization process is then used to encourage correct "associations", which make better use of the unlabeled data.
In \cite{Haitian_18_SemiSuperEntropyConstraints}, a method is proposed for combining several different semi-supervised learning techniques using Bayesian optimization.
In \cite{Papernot_2017_SemiSuperKnowledge} a semi-supervised framework that allows labeled training data privacy is proposed. In this framework, knowledge is transferred from teacher models to a student model in a semi-supervised manner, thereby precluding the student from gaining access to the labeled training data which is available to the teachers.

Two recent works that employ semi-supervised training techniques are \cite{Neal_18_SemiSupervised_Minimizing_Variance} and \cite{Abhishek_18_SemiSuper_GANs}.
In \cite{Neal_18_SemiSupervised_Minimizing_Variance} a semi-supervised deep kernel learning model is presented for regression tasks. 
In \cite{Abhishek_18_SemiSuper_GANs} a GAN based method is presented. It is proposed to estimate the tangent space to the learned data manifold using GANs, infer the relevant invariances and then inject these into the learned classifier during training.
In \cite{Oliver_2018_RealisticSSL} various semi-supervised learning algorithms are evaluated on real-world applications, yet no specific attention is paid to the transfer learning case and the effects of fine-tuning a pre-trained network.

Two works that focus on semi-supervised transfer learning are \cite{Gupta_2018_SemiSupervisSentiment} which examines semi-supervised transfer learning for sentiment classification, and \cite{Zhou_2018_SSL_TL} where semi-supervised transfer learning is examined for different training strategies and model choices. In the latter several observations regarding the application of existing semi-supervised methods in transfer learning settings are made. \\


\textbf{Information theory and machine learning.}
Information theory has lately been used to give theoretical insight into the intricacies of machine learning algorithms.
In \cite{Tishby99informationBotteleneck}, the Information Bottleneck framework has been presented. This framework formalizes the trade-off between algorithm sufficiency (fidelity) and complexity. It has been analyzed in various works such as \cite{Shamir_10_InformationBottleneckAnalysis} and \cite{Friedman_01_Multivariate_IB}. Following works \cite{Tishby15_information_deeplearning, Shwartz-ZivT17_blackbox} made the specific relation to deep learning, explicitly applying the principles of the information bottleneck to deep neural networks.
In \cite{Andrew_18_OnTheInformationBottleneck} several of the claims from \cite{Shwartz-ZivT17_blackbox} are examined and challenged.
In \cite{Marylou_18_MI_in_DL} useful methods for the computation of information theoretic quantities are proposed for several deep neural network models.

The closest work to ours is \cite{AchilleS18_InvarianceDisentanglement} in which an information theoretic approach is used in order to decompose the cross-entropy training loss of a machine learning algorithm into several separate terms. It is suggested that overfitting the training data is mathematically encapsulated in the mutual information between the training data labels and the learned model weights, i.e. this mutual information essentially represents the ability of a neural network to \emph{memorize} the training data \cite{GeneralizationError_18_Jakubovitz}.
Consequently, a regularizer that prevents overfitting is proposed, and initial results of its efficiency are presented.
However, unlike this work we propose a different decomposition of the cross-entropy \emph{test} loss and make the relation to semi-supervised transfer learning.
We propose a regularizer which leads to an improved semi-supervised transfer technique, and present experimental results that corroborate our theoretical analysis.

\section{The cross-entropy loss - an information theory perspective}
\label{sec:The Cross-Entropy Loss - an Information Theory Perspective}

Let $\mathcal{D} =\{ (x_i, y_i) \}_{i=1}^{N}$ be a training set with $N$ training samples that is used to train a learning algorithm with a set of weights $w$.
We assume that given $\mathcal{D}$ (a parameter of the model), the learning algorithm selects a specific hypothesis from the hypothesis class according to the distribution $p(w_\mathcal{D})$. In the case of a neural network, selecting the hypothesis is equivalent to training the network on the data.

We denote by $w_{\mathcal{D}}$ the model weights which were learned using the training set $\mathcal{D}$, and by $f(y|x,w_{\mathcal{D}})$ the learned classification function which given the weights $w_{\mathcal{D}}$ and a $D$-dimensional input $x \in \mathbb{R}^D$ computes the probability of the $K$-dimensional label $y \in \mathbb{R}^K$.
The learned classification function is tested on data drawn from the true underlying distribution $p(x,y)$.
Ideally, the learned classification function $f(y|x,w_{\mathcal{D}})$ would highly resemble the ground-truth classification $p(y|x)$, and similarly $f(x,y|w_{\mathcal{D}})$ would highly resemble $p(x,y)$.
With these notations, we turn to analyze the cross-entropy loss used predominantly in classification tasks. In our derivations we used several information theoretic measures which we present hereafter.

Let $X,Y$ be two random variables with respective probability density functions $p(x), p(y)$. The following are definitions of three information theoretic measures which are relevant to our derivations.

\begin{defn}[Mutual information]
The mutual information between $X$ and $Y$ is defined by
\begin{equation}
    I(X;Y) = \iint p(x,y) \log \left\{ \frac{p(x,y)}{p(x)p(y)} \right\} dx dy.
\end{equation}
\end{defn}

The Mutual information captures the dependence between two random variables. It is the Kullback-Leibler divergence between the joint distribution and the product of the marginal distributions.
The following is the Lautum information:
\begin{defn}[Lautum information]
The Lautum information between $X$ and $Y$ is
\begin{equation}
    L(X;Y) = \iint p(x)p(y) \log \left\{ \frac{p(x)p(y)}{p(x,y)} \right\} dx dy.
\end{equation}
\end{defn}

This measure is the Kullback-Leibler divergence between the product of the marginal distributions and the joint distribution.
Similar to the mutual information, the Lautum information is related to the dependence between two random variables. However, it has different properties than the mutual information, as outlined in \cite{Verdu_08_Lautum}. The last definition is of the differential entropy of a random variable.

\begin{defn}[Differential entropy]
The differential entropy of a random variable $X$ is defined by
\begin{equation}
    H(X) = - \int p(x) \log p(x) dx.
\end{equation}
\end{defn}

\noindent {\bf Main theoretical result.} Having these definitions, we present our main theoretical result which is given by the following theorem:

\begin{thm}
\label{thm:test_ce_loss}
For a classification task with ground-truth distribution $p(y|x)$, training set $\mathcal{D}$, learned weights $w_\mathcal{D}$ and learned classification function $f(y|x, w_\mathcal{D})$, the expected cross-entropy loss of a machine learning algorithm on the test distribution is equal to
\begin{equation}
    \label{eq:thm_test_ce_loss}
    \mathbb{E}_{w_{\mathcal{D}}} \left\{ KL( p(x,y) || f(x,y|w_{\mathcal{D}}) ) \right\} + H(y|x) - L(w_{\mathcal{D}};x).
\end{equation}
\end{thm}
Note that $KL$ signifies the Kullback-Leibler divergence and that we treat the training set $\mathcal{D}$ as a fixed parameter, whereas $w_{\mathcal{D}}$ and the examined test data $(x,y)$ are treated as random variables.
We refer the reader to Appendix~\ref{sec:Proof of Theorem 1} for the proof of Theorem~\ref{thm:test_ce_loss}.

In accordance with Theorem~\ref{thm:test_ce_loss}, the three terms that compose the expected cross-entropy test loss represent three different aspects of the loss of a learning algorithm performing a classification task:

\begin{itemize}
\item {\bf Classifier mismatch} $\mathbf{\mathbb{E}_{w_{\mathcal{D}}} KL \left( p(x,y) || f(x,y | w_{\mathcal{D}}) \right) }$: measures the deviation of the learned classification function's data distribution $f(x,y|w_{\mathcal{D}})$ from the true distribution of the data $p(x,y)$.
It is measured by the KL-divergence, which is averaged over all possible instances of $w$ parameterized by the training set $\mathcal{D}$. This term essentially measures the ability of the weights learned from $\mathcal{D}$ to represent the true distribution of the data.

\item {\bf Intrinsic Bayes error}
$\mathbf{H(y|x)}$: represents the inherent uncertainty of the labels given the data samples.

\item {\bf Lautum information}
between $w_{\mathcal{D}}$ and $x$, $\mathbf{L(w_\mathcal{D}; x) = \mathbb{E}_{w_{\mathcal{D}}} \{ KL( p(x) || p(x|w_{\mathcal{D}}) ) \}}$: represents the dependence between $w_{\mathcal{D}}$ and $x$. It essentially measures how much $p(x|w_{\mathcal{D}})$ deviates from $p(x)$ on average over the possible values of $w_{\mathcal{D}}$.
\end{itemize}

Our formulation suggests that a machine learning algorithm, which is trained relying on empirical risk minimization, implicitly aims at maximizing the Lautum information $L(w_\mathcal{D};x)$ in order to minimize the cross-entropy loss. At the same time, the algorithm aspires to minimize the KL-divergence between the ground-truth distribution of the data and the learned classification function.
The intrinsic Bayes error cannot be minimized and remains the inherent uncertainty of the task.
Namely, the formulation in \eqref{eq:thm_test_ce_loss} suggests that encouraging a larger Lautum information between the data samples and the learned model weights would be beneficial for reducing the model's test error on unseen data drawn from $p(x,y)$.

\section{Lautum information based semi-supervised transfer learning}
\label{sec:Lautum information based semi-supervised transfer learning}

We turn to show how we may apply our theory on the task of semi-supervised transfer learning.
In standard transfer learning, which consists of pre-transfer and post-transfer stages, a neural network is trained on a labeled source dataset and then fine-tuned on a smaller labeled target dataset.
In semi-supervised transfer learning, which we study here, we assume that an additional large set of unlabeled examples from the target distribution is available during training on the source data.

Semi-supervised transfer learning is highly beneficial in scenarios where the available target dataset is only partially annotated. Using the unlabeled part of this dataset, which is usually substantially bigger than the labeled part, has the potential of considerably improving the obtained performance. Thus, if this unlabeled part is a-priori available, then using it from the beginning of training can potentially improve the results.
For using the unlabeled samples of the target dataset during the pre-transfer training on the source dataset we leverage the formulation in \eqref{eq:thm_test_ce_loss}.
Considering its three terms, it is clear that by using unlabeled samples the classifier mismatch term cannot be minimized due to the lack of labels;
the intrinsic Bayes error is a characteristic of the task and cannot be minimized either; yet, the Lautum information does not depend on the labels and can therefore be calculated and maximized.

When the Lautum information is calculated between the model weights and data samples drawn from the target distribution, its maximization would encourage the learned weights to better relate to these samples, and by extension to better relate to the underlying probability distribution from which they were drawn.
Therefore, it is expected that an enlarged Lautum information will yield an improved performance on the target test set.
Accordingly, we aim at maximizing $L(w_{\mathcal{D}};x)$ during training. The pre-transfer maximization of the term $L(w_{\mathcal{D}};x)$, which is computed with samples drawn from the target distribution, would make the learned weights more inclined towards good performance on the target set right from the beginning. At the same time, the cross-entropy loss at this stage is calculated using labeled samples from the source dataset.
In the post-transfer stage, the cross-entropy loss is calculated using labeled samples from the target dataset, and therefore $L(w_{\mathcal{D}};x)$ is implicitly maximized during this stage.
We have empirically observed that explicitly maximizing the Lautum information between the unlabeled target samples and the model weights during post-transfer training (by imposing Lautum regularization) in addition to (or instead of) during pre-transfer training does not lead to improved results.

To summarize, our semi-supervised transfer learning approach optimizes two goals at the same time:
(i) minimizing the classifier mismatch $\mathbb{E}_{w_{\mathcal{D}}} \left\{ KL \left( p(x, y) || f(x, y | w_{\mathcal{D}}) \right)\right\}$, which is achieved using the labeled data both for the source and the target datasets during pre-transfer and post-transfer training respectively; and
(ii) maximizing the Lautum information $L(w_{\mathcal{D}};x)$, which is achieved explicitly using the unlabeled target data during pre-transfer training by imposing Lautum regularization, and in the post-transfer stage implicitly through the minimization of the cross-entropy loss which is evaluated on the labeled target data.
Figure~\ref{fig:Our_training_technique} summarizes our training scheme.

\subsection{Estimating the Lautum information}
\label{subsec: Estimating the Lautum information}

We are interested in using the Lautum information as a regularization term, which we henceforth refer to as "Lautum regularization".
Since computing the Lautum information between two random variables requires knowledge of their probability distribution functions (which are high-dimensional and hard to estimate), we assume that $w_{\mathcal{D}}$ and $x$ are jointly Gaussian with zero-mean. Even though this may seem like an arbitrary assumption, it nevertheless provides ease of computation and good experimental results as shown in Section~\ref{sec:Experiments}.

Since we only have one instance of the network weights at any specific point during training, we use the network features as a proxy for the network weights in the calculation of the Lautum information, instead of using the weights themselves. Namely, we use the network's output (its pre-softmax logits) when the input is $x$ as a proxy for the network weights $w_{\mathcal{D}}$. This way we have in every training iteration a number of samples equivalent to the size of our training mini-batch, instead of only one sample which would not allow any stable estimation to be made.

As shown in \cite{Verdu_08_Lautum}, the Lautum information between two jointly Gaussian random variables $(w,x)$ with covariance 
\begin{align}
\left[
\begin{array}{cc}
     \Sigma_{w} & \Sigma_{wx} \\
     \Sigma_{xw} & \Sigma_{x}
\end{array}
\right],
\end{align}
where $\Sigma_{x} \succ 0$ and $\Sigma_{w} \succ 0$, is given by
\begin{align}
\label{eq:Lautum_Gaussian}
\begin{split}
L(w;x) = & \log \left\{\det(I - \Sigma_{x}^{-1} \Sigma_{xw} \Sigma_{w}^{-1} \Sigma_{wx}) \right\} \\
& + 2 tr ((I - \Sigma_{x}^{-1} \Sigma_{xw} \Sigma_{w}^{-1} \Sigma_{wx})^{-1} - I).
\end{split}
\end{align}

The covariance matrix of our target dataset $\Sigma_{x}$ is evaluated once before training using the entire target training set, whereas $\Sigma_{w}, \Sigma_{wx}, \Sigma_{xw}$ are evaluated during training using the current mini-batch in every iteration.
All of these matrices are estimated using standard sample covariance estimation based on the current mini-batch in every iteration, e.g.
\begin{align}
\Sigma_{x} = \frac{1}{N_{batch}} \sum_{i=1}^{N_{batch}} (x_i - \mu_x) (x_i - \mu_x)^T,
\end{align}
and
\begin{align}
\Sigma_{xw} = \frac{1}{N_{batch}} \sum_{i=1}^{N_{batch}} (x_i - \mu_x) (w_i - \mu_w)^T,
\end{align}
where $$\mu_x = \frac{1}{N_{batch}} \sum_{i=1}^{N_{batch}} x_i, ~~~~~~ \mu_{w} = \frac{1}{N_{batch}} \sum_{i=1}^{N_{batch}} w_i$$ represent the sample mean values of $x$ and $w$ respectively (i.e. their average values in the current mini-batch) and $N_{batch}$ denotes the mini-batch size.
The dimensions of these matrices are $\Sigma_{x} \in \mathbb{R}^{D \times D}, \Sigma_{w} \in \mathbb{R}^{K \times K}, \Sigma_{xw} \in \mathbb{R}^{D \times K}, \Sigma_{wx} \in \mathbb{R}^{K \times D}$.
Note that $\Sigma_{wx} = \Sigma_{xw}^T$ hence only one of these matrices has to be calculated from the samples in every training iteration.

Since these are high-dimensional matrices, obtaining a numerically stable sample estimate would require a large amount of data, which would require the use of a very large mini-batch. This constraint poses both a hardware problem (since standard GPUs cannot fit mini-batches of thousands of examples) and a potential generalization degradation (as smaller mini-batches have been linked to improved generalization \cite{Keskar_2017_Large_Batch_Training}).
To overcome this issue, we use a standard exponentially decaying moving-average estimation of the three matrices $\Sigma_{w}, \Sigma_{wx}, \Sigma_{xw}$ in order to obtain numerical stability.
We denote by $\alpha$ the decay rate, and get the following update rule for the three covariance matrices in every training iteration:
\begin{align}
\label{eq:exp_sigma_sum}
    \Sigma_{(n)} = \alpha \Sigma_{(n-1)} + (1-\alpha) \Sigma_{batch},
\end{align}
where $n$ denotes the training iteration and $\Sigma_{batch}$ denotes the sample covariance matrix calculated using the current mini-batch. 
Using an exponentially decaying moving-average calculation is of particularly high importance for $\Sigma_{w}$, which is inverted to compute the Lautum regularization term.

\subsection{Training with Lautum regularization}
\label{subsec:Training with Lautum regularization}

Once the Lautum information has been estimated, our loss function for pre-transfer training is:
\begin{align}
\begin{split}
& Loss = \sum_{i=1}^{N} \sum_{k=1}^K -y_{ik}^{s} \log f_k(x_{i}^{s} | w_{\mathcal{D}}) - \lambda L(w_{\mathcal{D}}; x^{t}).
\end{split}
\end{align}
Note that the the cross-entropy loss is calculated using labeled samples from the source training set (which we denote by the $s$ superscript) whereas the Lautum regularization term is calculated using unlabeled samples from the target training set (which we denote by the $t$ superscript).
Also note that $y_i$ represents the ground truth label of the sample $x_i$; $f(x_{i} | w_{\mathcal{D}})$ represents the network's estimated post softmax label for that sample; and $L(w_{\mathcal{D}};x)$ is calculated as detailed in Section~\ref{subsec: Estimating the Lautum information}.
We emphasize that the Lautum regularization term is subtracted and not added to the cross-entropy loss since we aim at \emph{maximizing} the Lautum information during training.
Our loss function for post-transfer training consists of a standard cross-entropy loss:
\begin{align}
\begin{split}
& Loss = \sum_{i=1}^{N} \sum_{k=1}^K -y_{ik}^{t} \log f_k(x_{i}^{t} | w_{\mathcal{D}}).
\end{split}
\end{align}
Note that at this stage the cross-entropy loss, which is calculated using labeled target samples, inherently includes the Lautum term of the target data (see Theorem~\ref{thm:test_ce_loss}).

\section{Experiments}
\label{sec:Experiments}
In order to demonstrate the advantages of semi-supervised transfer learning with Lautum regularization we perform several experiments on image classification tasks using deep neural networks (though our theoretical derivations also apply to other machine learning algorithms).

\subsection{Experimental setup}
\label{subsec:Experimental setup}
We train deep neural networks and perform transfer learning from the original source dataset to the target dataset.
In our experiments we use the original labeled source training set as is and split the target training set into two parts. The first part is very small and contains labeled samples, whereas the second part consists of the remainder of the target training set and contains unlabeled samples only (the labels are discarded).
The performance is evaluated by the post transfer accuracy on the target test set.

We examine five different methods of transfer learning:
(1) standard supervised transfer which uses the labeled samples only.
(2) Temporal Ensembling semi-supervised learning as outlined in \cite{Laine_18_TemporalEnsembling}, applied in a transfer learning setting. Temporal Ensembling is applied in the post-transfer training stage.
(3) Mk-MMD \cite{Mk_MMD}, which is based on 19 different Gaussian kernels with different standard deviations. Mk-MMD is applied in the pre-transfer training stage.
(4) Lautum regularization - our technique as described in Section~\ref{sec:Lautum information based semi-supervised transfer learning}.
(5) Both Temporal Ensembling and Lautum regularization. Note that Temporal Ensembling is applied in the post-transfer training stage, whereas Lautum regularization is applied in the pre-transfer training stage.

As presented in Section~\ref{subsec:Training with Lautum regularization}, our training consists of two stages. First, we train the network using our fully labeled source training set while using Lautum regularization with the unlabeled samples from our target training set.
We use the same mini-batch size both for the calculation of the cross-entropy loss (using labeled source samples) and for the computation of the Lautum regularization term (using unlabeled target samples).
We also use an exponentially decaying moving average to obtain numerical stability in the estimation of the the covariance matrices $\Sigma_w, \Sigma_{wx}, \Sigma_{xw}$.
The matrix $\Sigma_x$ is calculated once before training and remains constant all throughout it.

Second, we perform a transfer to the target set by training (fine-tuning) the entire network using the labeled samples from the target training set, where the mini-batch size remains the same as before. As in \cite{Zhou_2018_SSL_TL}, we fine-tune the entire network since this best fits the settings of semi-supervised learning. As mentioned above, we do not apply the Lautum regularization at this stage as we empirically found that it does not improve the results.

We perform our experiments on the MNIST and CIFAR-10 datasets.
For MNIST we examine the transfer to the notMNIST dataset, which consists of 10 classes representing the letters A-J.
The notMNIST dataset is similar to the MNIST dataset in its grayscale styling and image size, yet it differs in content.
For CIFAR-10 we examine the transfer to 10 specific classes of the CIFAR-100 dataset (specifically, classes 0, 10, 20,...,90 of CIFAR-100, which we reclassified as classes 0, 1, 2,...,9 respectively).
These CIFAR-100 classes are different than the corresponding CIFAR-10 ones in content.
For example, class 0 in CIFAR-10 represents airplanes whereas class 0 in CIFAR-100 represents beavers etc.
Both in the MNIST $\rightarrow$ notMNIST case and the CIFAR-10 $\rightarrow$ CIFAR-100 (10 classes) case we used the same CNN as in \cite{Laine_18_TemporalEnsembling}. The architecture of the network is illustrated in Appendix~\ref{sec:The CNN architecture used in the experiments}.

\subsection{MNIST to notMNIST results}
\label{subsec:MNIST to notMNIST results}
In order for the input images to fit the network's input we resized the MNIST and notMNIST images to $32$x$32$ pixels and transformed each of them to RGB format.
Training was done using an Adam optimizer \cite{Kingma15Adam} and a mini-batch size of 50 inputs. With this network and using standard supervised training on the entire MNIST dataset we obtained a test accuracy of $99.01\%$ on MNIST.

For each of the five transfer learning methods we examined three different splits of the notMNIST training dataset which consists of 200,000 samples to an unlabeled part and a labeled part: (1) unlabeled part of 199,950 samples and a labeled part of 50 samples; (2) unlabeled part of 199,900 samples and a labeled part of 100 samples; (3) unlabeled part of 199,800 samples and a labeled part of 200 samples. All three options use very few labeled samples in order to fairly represent realistic semi-supervised learning scenarios - in all three options 99.9\% or more of the training data is unlabeled.
We used a decay rate of $\alpha = 0.999$ for the exponentially decaying moving average estimation of the 3 covariance matrices $\Sigma_w, \Sigma_{wx}, \Sigma_{xw}$, and a different value of $\lambda$ (which controls the weight of the Lautum regularization) in each scenario which we found to provide a good balance between the cross-entropy loss and the Lautum regularization.

Using the settings outlined above we obtained the results shown in Table~\ref{table:MkMMD_TE_Lautum_Comparison_MNIST} for the MNIST $\rightarrow$ notMNIST case.
The advantage of using Lautum regularization is evident from the results, as it outperforms the other compared methods in all the examined target training set splits.
In general, the Temporal Ensembling method by itself does not yield very competitive results compared to standard transfer learning.
However, the combination of Lautum regularization and Temporal Ensembling provides the best target test set accuracy over all the examined target training set splits.



\begin{table}[ht!]
\begin{center}
\begin{tabular}{|c|c|c|c|}
 \hline
   \textbf{Method} &
   \textbf{Source} $\rightarrow$ \textbf{Target} &
   \textbf{\# labeled} &
   \textbf{Accuracy} \\ [0.5ex]
   \hline \hline
   Standard & MNIST / notMNIST & 50 & 34.02\%  \\
   TE & MNIST / notMNIST & 50 & 37.28\% \\
   Mk-MMD & MNIST / notMNIST & 50 & 46.72\% \\
   Lautum & MNIST / notMNIST & 50 & 47.96\% \\
   Lautum + TE & MNIST / notMNIST & 50 & 66.91\% \\
   \hline
   Standard & MNIST / notMNIST & 100 & 57.58\% \\
   TE & MNIST / notMNIST & 100 & 61.45\% \\
   Mk-MMD & MNIST / notMNIST & 100 & 63.32\% \\
   Lautum & MNIST / notMNIST & 100 & 65.21\% \\
   Lautum + TE & MNIST / notMNIST & 100 & 77.32\% \\
   \hline
   Standard & MNIST / notMNIST & 200 & 67.78\% \\
   TE & MNIST / notMNIST & 200 & 74.87\% \\
   Mk-MMD & MNIST / notMNIST & 200 & 80.35\% \\
   Lautum & MNIST / notMNIST & 200 & 83.77\% \\
   Lautum + TE & MNIST / notMNIST & 200 & 85.25\% \\
  \hline
\end{tabular}
\end{center}
\caption{target test set accuracy comparison between standard transfer learning, Temporal Ensembling (TE), Mk-MMD, Lautum regularization and both TE and Lautum regularization for different amounts of labeled training target samples, MNIST $\rightarrow$ notMNIST.}
\label{table:MkMMD_TE_Lautum_Comparison_MNIST}
\end{table}

\subsection{CIFAR-10 to CIFAR-100 (10 classes) results}
\label{subsec:CIFAR-10 to CIFAR-100 (10 classes) results}

In the CIFAR-10 $\rightarrow$ CIFAR-100 (10 classes) case training was done using an Adam optimizer \cite{Kingma15Adam} and a mini-batch size of 100 inputs. With this network and using standard supervised training on the entire CIFAR-10 dataset we obtained a test accuracy of $85.09\%$ on CIFAR-10.

Our target set consists of 10 classes of the CIFAR-100 dataset. Accordingly, our training target set consists of 5,000 samples and our test target set consists of 1,000 samples.
We examined the same five transfer learning techniques as in the MNIST $\rightarrow$ notMNIST case, where for each we examined three different splits of the CIFAR-100 (10 classes) training dataset to an unlabeled part and a labeled part: (1) unlabeled part of 4,900 samples and a labeled part of 100 samples; (2) unlabeled part of 4,800 samples and a labeled part of 200 samples; (3) unlabeled part of 4,500 samples and a labeled part of 500 samples. All three options use a small number of labeled samples in order to fairly represent realistic semi-supervised learning scenarios - in all three options 90\% or more of the data is unlabeled.
We used a decay rate of $\alpha = 0.999$ for the exponentially decaying moving average estimation of the 3 covariance matrices $\Sigma_w, \Sigma_{wx}, \Sigma_{xw}$, and a different value of $\lambda$ (which controls the weight of the Lautum regularization) in each scenario which we found to provide a good balance between the cross-entropy loss and the Lautum regularization.

Using the settings outlined above we obtained the results shown in Table~\ref{table:MkMMD_TE_Lautum_Comparison_CIFAR} for the CIFAR-10 $\rightarrow$ CIFAR-100 (10 classes) case.
It is evident from the results that in the CIFAR-10 $\rightarrow$ CIFAR-100 (10 classes) case as well, using Lautum regularization improves the post-transfer performance on the target test set and outperforms Temporal Ensembling and Mk-MMD.
Note, however, that in the case of 200 and 500 labeled samples, applying Lautum regularization only, without Temporal Ensembling, slightly outperforms the combination of the two methods.

\begin{table}[ht!]
\begin{center}
\begin{tabular}{|c|c|c|c|}
 \hline
   \textbf{Method} &
   \textbf{Source} $\rightarrow$ \textbf{Target} &
   \textbf{\# labeled} &
   \textbf{Accuracy} \\ [0.5ex]
   \hline \hline
   Standard & CIFAR-10 / 100 & 100 & 39.90\% \\
   TE & CIFAR-10 / 100 & 100 & 42.20\% \\
   Mk-MMD & CIFAR-10 / 100 & 100 & 45.30\% \\
   Lautum & CIFAR-10 / 100 & 100 & 46.70\% \\
   Lautum + TE & CIFAR-10 / 100 & 100 & 46.90\% \\
   \hline
   Standard & CIFAR-10 / 100 & 200 & 52.80\%  \\
   TE & CIFAR-10 / 100 & 200 & 54.60\% \\
   Mk-MMD & CIFAR-10 / 100 & 200 & 59.30\% \\
   Lautum & CIFAR-10 / 100 & 200 & 60.90\% \\
   Lautum + TE & CIFAR-10 / 100 & 200 & 60.40\% \\
   \hline
   Standard & CIFAR-10 / 100 & 500 & 64.50\% \\
   TE & CIFAR-10 / 100 & 500 & 66.50\% \\
   Mk-MMD & CIFAR-10 / 100 & 500 & 68.00\% \\
   Lautum & CIFAR-10 / 100 & 500 & 70.80\% \\
   Lautum + TE & CIFAR-10 / 100 & 500 & 70.30\% \\
  \hline
\end{tabular}
\end{center}
\caption{target test set accuracy comparison between standard transfer learning, Temporal Ensembling (TE), Mk-MMD, Lautum regularization and both TE and Lautum regularization for different amounts of labeled training target samples, CIFAR-10 $\rightarrow$ CIFAR-100 (10 classes).}
\label{table:MkMMD_TE_Lautum_Comparison_CIFAR}
\end{table}

\section{Conclusions}
\label{sec:Conclusions}

We proposed a new semi-supervised transfer learning approach for machine learning algorithms that are trained using the cross-entropy loss. Our approach is backed by information theoretic derivations and exemplifies how one can make good use of unlabeled samples along with just a few labeled samples to improve performance on the target dataset.
Our approach relies on the maximization of the Lautum information between unlabeled samples from the target set and an algorithm's learned features by using the Lautum information as a regularization term. As shown, the maximization of the Lautum information minimizes the cross-entropy test loss on the target set and thereby improves performance as indicated by our experimental results.
We have also shown that our approach surpasses the performance of prominent state-of-the-art semi-supervised learning techniques in a transfer learning setting.

Future work will focus on alternative approximations of the Lautum information which could potentially yield better performance or reduce the additional computational overhead it introduces.
In addition, our formulation has the potential to be applied in other tasks as well, such as multi-task learning or domain adaptation.
Incorporating techniques to mitigate the effects of training using an imbalanced dataset could also be of interest.
We defer these directions to future research. \\

\appendices

\section{Proof of Theorem 1}
\label{sec:Proof of Theorem 1}

Let us reiterate Theorem 1 before formally proving it.

\noindent \textbf{Theorem 1}
\emph{For a classification task with ground-truth distribution $p(y|x)$, training set $\mathcal{D}$, learned weights $w_\mathcal{D}$ and learned classification function $f(y|x, w_\mathcal{D})$, the expected cross-entropy loss of a machine learning algorithm on the test distribution is equal to}
\begin{equation}
    \label{eq:CE_test_loss_expression_appendix}
    \mathbb{E}_{w_{\mathcal{D}}} \left\{ KL( p(x,y) || f(x,y|w_{\mathcal{D}}) ) \right\} + H(y|x) - L(w_{\mathcal{D}};x).
\end{equation}

\noindent \textbf{Proof.}
The expected cross-entropy loss of the learned classification function $f(y|x, w_\mathcal{D})$ on the test distribution $p(x,y)$ is given by
\begin{equation}
    \label{eq:Test_CE_appendix}
    \mathbb{E}_{(x,y) \sim p(x,y)} \mathbb{E}_{w \sim p(w_\mathcal{D})} \{-\log f(y|x,w_\mathcal{D})\}.
\end{equation}

Explicitly, \eqref{eq:Test_CE_appendix} can be written as
\footnote{Since the values of $y$ are discrete it is more accurate to sum instead of integrate over them. Yet, for the simplicity of the proof we present the derivations using integration.}
\begin{equation}
    \label{eq:full_expression_appendix}
    - \iiint p(x,y) p(w_\mathcal{D}) \log f(y|x,w_\mathcal{D}) \hspace{0.05 cm} dx \hspace{0.05 cm} dy \hspace{0.05 cm} dw_\mathcal{D}.
\end{equation}

To compare the learned classifier with the true classification of the data we develop \eqref{eq:full_expression_appendix} further as follows:
\begin{equation}
    = - \iiint p(x,y) p(w_\mathcal{D}) \log \left\{ \frac{f(y|x, w_\mathcal{D})}{p(y|x, w_\mathcal{D})} p(y|x, w_\mathcal{D}) \right\} \hspace{0.05 cm} dx \hspace{0.05 cm} dy \hspace{0.05 cm} dw_\mathcal{D}.
\end{equation}

Using standard logarithm arithmetic we get the following expression:
\begin{equation}
    \label{eq:Two_expressions}
    \begin{split}
     = & \underbrace{- \iiint p(x,y) p(w_\mathcal{D}) \log \left\{ \frac{f(y|x, w_\mathcal{D})}{p(y|x, w_\mathcal{D})} \right\} \hspace{0.05 cm} dx \hspace{0.05 cm} dy \hspace{0.05 cm} dw_\mathcal{D}}_{(\star)} \\
    & \underbrace{- \iiint p(x,y) p(w_\mathcal{D}) \log p(y|x, w_\mathcal{D}) \hspace{0.05 cm} dx \hspace{0.05 cm} dy \hspace{0.05 cm} dw_\mathcal{D}}_{(\star \star)}.
    \end{split}
\end{equation}

We separate the derivations of the two terms in \eqref{eq:Two_expressions}.
First, we develop the term $(\star \star)$ further:
\begin{equation}
    \begin{split}
    (\star \star) = &
    - \iiint p(x,y) p(w_{\mathcal{D}}) \log p(y|x,w_{\mathcal{D}}) \hspace{0.05 cm} dx \hspace{0.05 cm} dy \hspace{0.05 cm} dw_\mathcal{D} \\
    = & - \iiint p(x,y)p(w_{\mathcal{D}}) \log \left\{ \frac{p(x,y,w_{\mathcal{D}})}{p(x,w_{\mathcal{D}})} \hspace{0.05 cm} \right\} dx \hspace{0.05 cm} dy \hspace{0.05 cm} dw_\mathcal{D}.
    \end{split}
\end{equation}

Using logarithm arithmetic and adding and subtracting terms we get:
\begin{equation}
\begin{split}
\label{eq:star_4_terms_appendix}
(\star \star) = 
& - \iiint p(x,y) p(w_{\mathcal{D}}) \log \left\{ \frac{p(x, y, w_{\mathcal{D}})}
{p(x,y) p(w_{\mathcal{D}})} \right\} dx \hspace{0.05 cm} dy \hspace{0.05 cm} dw_\mathcal{D}
\\
& - \iiint p(x,y) p(w_{\mathcal{D}}) \log \left\{ p(x,y) p(w_{\mathcal{D}})
\right\} dx \hspace{0.05 cm} dy \hspace{0.05 cm} dw_\mathcal{D}
\\
& - \iiint p(x,y) p(w_{\mathcal{D}}) \log \left\{ \frac{p(x) p(w_{\mathcal{D}})}
{p(x, w_{\mathcal{D}})} \right\} dx \hspace{0.05 cm} dy \hspace{0.05 cm} dw_\mathcal{D}
\\
& + \iiint p(x,y) p(w_{\mathcal{D}}) \log \left\{ p(x) p(w_{\mathcal{D}}) \right\} dx \hspace{0.05 cm} dy \hspace{0.05 cm} dw_\mathcal{D}.
\end{split}
\end{equation}

Using the law of total probability along with the definitions of the differential entropy and the Lautum information we can reformulate \eqref{eq:star_4_terms_appendix} as follows:
\begin{equation}
\begin{split}
    (\star \star)
    & = L(w_{\mathcal{D}}; (x,y)) \\
    & + H(w_{\mathcal{D}}) + H(x,y) \\
    & - L(w_{\mathcal{D}};x) \\
    & - H(x) - H(w_{\mathcal{D}}).
    \end{split}
\end{equation}

Since $H(y|x) = H(x,y) - H(x)$ we get that:
\begin{equation}
    \label{eq:star_final_exp}
    (\star \star) = L(w_{\mathcal{D}};(x,y)) + H(y|x) - L(w_{\mathcal{D}};x).
\end{equation}

We next analyze the expression of $(\star)$:
\begin{equation}
    \begin{split}
    (\star) = &
    - \iiint p(x,y) p(w_{\mathcal{D}}) \log \left\{ \frac{f(y|x,w_{\mathcal{D}})}{p(y|x,w_{\mathcal{D}})} \right\} \hspace{0.05 cm} dx \hspace{0.05 cm} dy \hspace{0.05 cm} dw_\mathcal{D}.
    \end{split}
\end{equation}

Within the $\log$ operation we multiply and divide by the term $\frac{p(x,y)p(w_{\mathcal{D}})}{p(x,w_{\mathcal{D}})}$ and get:
\begin{equation}
    \begin{split}
    & (\star) = \iiint p(x,y) p(w_{\mathcal{D}}) \cdot \\
    & \log \left\{
    \frac{p(x,y)p(w_{\mathcal{D}})}{f(y | x, w_{\mathcal{D}}) p(x,w_{\mathcal{D}})}
    \cdot 
    \frac{p(y | x, w_{\mathcal{D}}) p(x,w_{\mathcal{D}})}{p(x,y)p(w_{\mathcal{D}})}
    \right\}
    dx \hspace{0.05 cm} dy \hspace{0.05 cm} dw_\mathcal{D}.
    \end{split}
\end{equation}

Since $f(y|x,w_{\mathcal{D}})$ is the learned classifier which outputs the probability of the label $y$ for an input $x$ given the model weights $w_{\mathcal{D}}$, without the labels it has no affect on the joint distribution of the weights and inputs, i.e. $f(x,w_{\mathcal{D}}) = p(x,w_{\mathcal{D}})$, $f(w_{\mathcal{D}}) = p(w_{\mathcal{D}})$. Accordingly,
\begin{equation}
\begin{split}
    (\star) = & \iiint p(x,y) p(w_{\mathcal{D}}) \log \left\{ \frac{p(x,y)p(w_{\mathcal{D}})}{f(x,y,w_{\mathcal{D}})} \right\} dx \hspace{0.05 cm} dy \hspace{0.05 cm} dw_\mathcal{D} \\
    - & \iiint p(x,y) p(w_{\mathcal{D}}) \log \left\{ \frac{p(x,y)p(w_{\mathcal{D}})}{p(x,y,w_{\mathcal{D}})} \right\} dx \hspace{0.05 cm} dy \hspace{0.05 cm} dw_\mathcal{D}.
\end{split}
\end{equation}

Since $f(x,y|w_{\mathcal{D}}) = \frac{f(x,y,w_{\mathcal{D}})}{p(w_{\mathcal{D}})}$ we get that:
\begin{equation}
\begin{split}
    (\star) = & \iiint p(x,y) p(w_{\mathcal{D}}) \log \left\{ \frac{p(x,y)}{f(x,y|w_{\mathcal{D}})} \right\} dx \hspace{0.05 cm} dy \hspace{0.05 cm} dw_\mathcal{D} \\
    - & \iiint p(x,y) p(w_{\mathcal{D}}) \log \left\{ \frac{p(x,y)p(w_{\mathcal{D}})}{p(x,y,w_{\mathcal{D}})} \right\} dx \hspace{0.05 cm} dy \hspace{0.05 cm} dw_\mathcal{D}.
\end{split}
\end{equation}

In the first term we have the expectation over $w_\mathcal{D}$ and so:
\begin{equation}
\begin{split}
    (\star) = &  \mathbb{E}_{w_{\mathcal{D}}} \left\{ \iint p(x,y) \log \left\{ \frac{p(x,y)}{f(x,y|w_{\mathcal{D}})} \right\} dx \hspace{0.05 cm} dy \right\} \\
    - & \iiint p(x,y) p(w_{\mathcal{D}}) \log \left\{ \frac{p(x,y)p(w_{\mathcal{D}})}{p(x,y,w_{\mathcal{D}})} \right\} dx \hspace{0.05 cm} dy \hspace{0.05 cm} dw_\mathcal{D}.
\end{split}
\end{equation}

We get that the first term is the expectation over $w_\mathcal{D}$ of the KL-divergence between $p(x,y)$ and $f(x,y|w_{\mathcal{D}})$, whereas the second term is the negative Lautum information between $(x,y)$ and $w_{\mathcal{D}}$:
\begin{equation}
\begin{split}
    \label{eq:star_star_final_exp}
    (\star) = \mathbb{E}_{w_{\mathcal{D}}} \left\{ KL( p(x,y) || f(x,y|w_{\mathcal{D}}) ) \right\}
    - L(w_{\mathcal{D}};(x,y)).
\end{split}
\end{equation}

Plugging the expressions we got for $(\star \star)$ from \eqref{eq:star_final_exp} and for $(\star)$ from \eqref{eq:star_star_final_exp} into \eqref{eq:Two_expressions} we obtain the expression in \eqref{eq:CE_test_loss_expression_appendix}:
\begin{equation}
    \label{eq:final_term}
    \begin{split}
    & (\star) \hspace{0.05cm} + \hspace{0.05cm} (\star \star) =
    \underbrace{\mathbb{E}_{w_{\mathcal{D}}} \left\{ KL( p(x,y) || f(x,y|w_{\mathcal{D}}) ) \right\} - L(w_{\mathcal{D}};(x,y))}_{(\star)} \\
    & + \underbrace{L(w_{\mathcal{D}};(x,y)) + H(y|x) - L(w_{\mathcal{D}};x)}_{(\star \star)} \\
    & = \mathbb{E}_{w_{\mathcal{D}}} \left\{ KL( p(x,y) || f(x,y|w_{\mathcal{D}}) ) \right\} + H(y|x) - L(w_{\mathcal{D}};x).
    \end{split}
\end{equation}

\section{The CNN architecture used in the experiments}
\label{sec:The CNN architecture used in the experiments}

\begin{figure}
    \centering
    \includegraphics[width=\columnwidth]{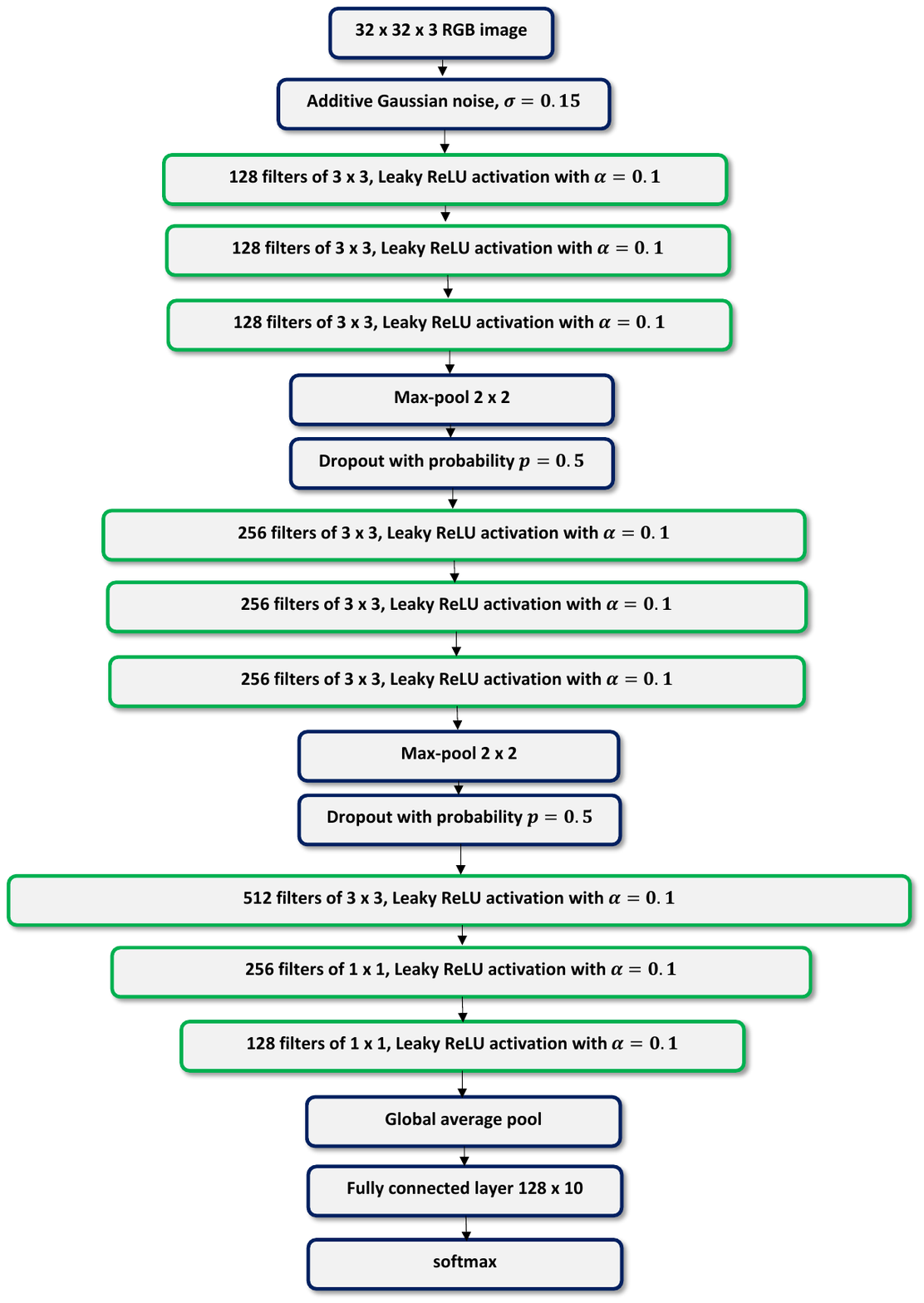}
    \caption{The network architecture used in our experiments.} \label{fig:Network_architecture}
\end{figure}

Both in the MNIST $\rightarrow$ notMNIST case and the CIFAR-10 $\rightarrow$ CIFAR-100 (10 classes) case we used the same CNN as in \cite{Laine_18_TemporalEnsembling}. The architecture is illustrated in Figure~\ref{fig:Network_architecture}.

\section*{Acknowledgment}
This work was supported by the ERC-StG SPADE grant.

\ifCLASSOPTIONcaptionsoff
  \newpage
\fi

\bibliographystyle{IEEEtran}
\bibliography{egbib_for_arxiv_v2}

\begin{thebibliography}{10}
\providecommand{\url}[1]{#1}
\csname url@samestyle\endcsname
\providecommand{\newblock}{\relax}
\providecommand{\bibinfo}[2]{#2}
\providecommand{\BIBentrySTDinterwordspacing}{\spaceskip=0pt\relax}
\providecommand{\BIBentryALTinterwordstretchfactor}{4}
\providecommand{\BIBentryALTinterwordspacing}{\spaceskip=\fontdimen2\font plus
\BIBentryALTinterwordstretchfactor\fontdimen3\font minus
  \fontdimen4\font\relax}
\providecommand{\BIBforeignlanguage}[2]{{%
\expandafter\ifx\csname l@#1\endcsname\relax
\typeout{** WARNING: IEEEtran.bst: No hyphenation pattern has been}%
\typeout{** loaded for the language `#1'. Using the pattern for}%
\typeout{** the default language instead.}%
\else
\language=\csname l@#1\endcsname
\fi
#2}}
\providecommand{\BIBdecl}{\relax}
\BIBdecl

\bibitem{Goodfellow16Deep}
I.~Goodfellow, Y.~Bengio, and A.~Courville, \emph{Deep Learning}.\hskip 1em
  plus 0.5em minus 0.4em\relax MIT Press, 2016.

\bibitem{Donahue_2014_Decaf}
\BIBentryALTinterwordspacing
J.~Donahue, Y.~Jia, O.~Vinyals, J.~Hoffman, N.~Zhang, E.~Tzeng, and T.~Darrell,
  ``Decaf: A deep convolutional activation feature for generic visual
  recognition,'' in \emph{Proceedings of the 31st International Conference on
  Machine Learning}, ser. Proceedings of Machine Learning Research, E.~P. Xing
  and T.~Jebara, Eds., vol.~32, no.~1.\hskip 1em plus 0.5em minus 0.4em\relax
  Bejing, China: PMLR, 22--24 Jun 2014, pp. 647--655. [Online]. Available:
  \url{http://proceedings.mlr.press/v32/donahue14.html}
\BIBentrySTDinterwordspacing

\bibitem{Xuhong_2018_Explicit_TL}
\BIBentryALTinterwordspacing
X.~LI, Y.~Grandvalet, and F.~Davoine, ``Explicit inductive bias for transfer
  learning with convolutional networks,'' in \emph{Proceedings of the 35th
  International Conference on Machine Learning}, ser. Proceedings of Machine
  Learning Research, J.~Dy and A.~Krause, Eds., vol.~80.\hskip 1em plus 0.5em
  minus 0.4em\relax StockholmsmÃ¤ssan, Stockholm Sweden: PMLR, 10--15 Jul
  2018, pp. 2825--2834. [Online]. Available:
  \url{http://proceedings.mlr.press/v80/li18a.html}
\BIBentrySTDinterwordspacing

\bibitem{AchilleS18_InvarianceDisentanglement}
\BIBentryALTinterwordspacing
A.~Achille and S.~Soatto, ``Emergence of invariance and disentanglement in deep
  representations,'' \emph{Journal of Machine Learning Research}, vol.~19,
  2018. [Online]. Available:
  \url{http://jmlr.org/papers/v19/papers/v19/17-646.html}
\BIBentrySTDinterwordspacing

\bibitem{Verdu_08_Lautum}
D.~P. Palomar and S.~Verdu, ``Lautum information,'' \emph{IEEE Trans. Inform.
  Theory}, vol.~54, no.~3, pp. 964--975, March 2008.

\bibitem{Laine_18_TemporalEnsembling}
S.~Laine and T.~Aila, ``Temporal ensembling for semi-supervised learning,'' in
  \emph{ICLR}, 2017.

\bibitem{Mk_MMD}
\BIBentryALTinterwordspacing
A.~Gretton, D.~Sejdinovic, H.~Strathmann, S.~Balakrishnan, M.~Pontil,
  K.~Fukumizu, and B.~K. Sriperumbudur, ``Optimal kernel choice for large-scale
  two-sample tests,'' in \emph{Advances in Neural Information Processing
  Systems 25}, F.~Pereira, C.~J.~C. Burges, L.~Bottou, and K.~Q. Weinberger,
  Eds.\hskip 1em plus 0.5em minus 0.4em\relax Curran Associates, Inc., 2012,
  pp. 1205--1213. [Online]. Available:
  \url{http://papers.nips.cc/paper/4727-optimal-kernel-choice-for-large-scale-two-sample-tests.pdf}
\BIBentrySTDinterwordspacing

\bibitem{JialinPan_10_TransferSurvey}
\BIBentryALTinterwordspacing
S.~J. Pan and Q.~Yang, ``A survey on transfer learning,'' \emph{{IEEE} Trans.
  Knowl. Data Eng.}, vol.~22, no.~10, pp. 1345--1359, 2010. [Online].
  Available: \url{https://doi.org/10.1109/TKDE.2009.191}
\BIBentrySTDinterwordspacing

\bibitem{Tan_2018_DL_Transfer_Survey}
C.~Tan, F.~Sun, T.~Kong, W.~Zhang, C.~Yang, and C.~Liu, ``A survey on deep
  transfer learning,'' in \emph{ICANN}, 2018.

\bibitem{Jason_2014_HowTransferable}
\BIBentryALTinterwordspacing
J.~Yosinski, J.~Clune, Y.~Bengio, and H.~Lipson, ``How transferable are
  features in deep neural networks?'' in \emph{Advances in Neural Information
  Processing Systems 27}, Z.~Ghahramani, M.~Welling, C.~Cortes, N.~D. Lawrence,
  and K.~Q. Weinberger, Eds.\hskip 1em plus 0.5em minus 0.4em\relax Curran
  Associates, Inc., 2014, pp. 3320--3328. [Online]. Available:
  \url{http://papers.nips.cc/paper/5347-how-transferable-are-features-in-deep-neural-networks.pdf}
\BIBentrySTDinterwordspacing

\bibitem{Lampinen_19_TransferLinear}
A.~K. Lampinen and S.~Ganguli, ``An analytic theory of generalization dynamics
  and transfer learning in deep linear networks,'' in \emph{ICLR}, 2019.

\bibitem{Xuezhi_2014_Flexible_Transfer}
\BIBentryALTinterwordspacing
X.~Wang and J.~Schneider, ``Flexible transfer learning under support and model
  shift,'' in \emph{Advances in Neural Information Processing Systems 27},
  Z.~Ghahramani, M.~Welling, C.~Cortes, N.~D. Lawrence, and K.~Q. Weinberger,
  Eds.\hskip 1em plus 0.5em minus 0.4em\relax Curran Associates, Inc., 2014,
  pp. 1898--1906. [Online]. Available:
  \url{http://papers.nips.cc/paper/5632-flexible-transfer-learning-under-support-and-model-shift.pdf}
\BIBentrySTDinterwordspacing

\bibitem{Wei_18_Learning_to_transfer}
\BIBentryALTinterwordspacing
Y.~Wei, Y.~Zhang, J.~Huang, and Q.~Yang, ``Transfer learning via learning to
  transfer,'' in \emph{Proceedings of the 35th International Conference on
  Machine Learning}, ser. Proceedings of Machine Learning Research, J.~Dy and
  A.~Krause, Eds., vol.~80.\hskip 1em plus 0.5em minus 0.4em\relax
  Stockholmsmässan, Stockholm Sweden: PMLR, 10--15 Jul 2018, pp. 5085--5094.
  [Online]. Available: \url{http://proceedings.mlr.press/v80/wei18a.html}
\BIBentrySTDinterwordspacing

\bibitem{Zamir_2018_Taskonomy}
A.~R. Zamir, A.~Sax, W.~Shen, L.~J. Guibas, J.~Malik, and S.~Savarese,
  ``Taskonomy: Disentangling task transfer learning,'' in \emph{The IEEE
  Conference on Computer Vision and Pattern Recognition (CVPR)}, June 2018.

\bibitem{Zhu_06_SemiSupervisedSurvey}
X.~Zhu, ``Semi-supervised learning literature survey,'' in \emph{University of
  Wisconsin-–Madison, Tech. Rep. 1530}, 2006.

\bibitem{Yves_05_SemiSupervised_Entropy}
Y.~Grandvalet and Y.~Bengio, ``Semi-supervised learning by entropy
  minimization,'' in \emph{Advances in Neural Information Processing Systems
  17}, L.~K. Saul, Y.~Weiss, and L.~Bottou, Eds.\hskip 1em plus 0.5em minus
  0.4em\relax MIT Press, 2005, pp. 529--536.

\bibitem{HausserMC_17_LearningByAssociation}
\BIBentryALTinterwordspacing
P.~H{\"{a}}usser, A.~Mordvintsev, and D.~Cremers, ``Learning by association -
  {A} versatile semi-supervised training method for neural networks,'' in
  \emph{2017 {IEEE} Conference on Computer Vision and Pattern Recognition,
  {CVPR} 2017, Honolulu, HI, USA, July 21-26, 2017}, 2017, pp. 626--635.
  [Online]. Available: \url{https://doi.org/10.1109/CVPR.2017.74}
\BIBentrySTDinterwordspacing

\bibitem{Haitian_18_SemiSuperEntropyConstraints}
H.~Sun, W.~W. Cohen, and L.~Bing, ``Semi-supervised learning with declaratively
  specified entropy constraints,'' in \emph{Advances in Neural Information
  Processing Systems 31}, S.~Bengio, H.~Wallach, H.~Larochelle, K.~Grauman,
  N.~Cesa-Bianchi, and R.~Garnett, Eds.\hskip 1em plus 0.5em minus 0.4em\relax
  Curran Associates, Inc., 2018, pp. 4430--4440.

\bibitem{Papernot_2017_SemiSuperKnowledge}
N.~Papernot, M.~Abadi, Úlfar Erlingsson, I.~Goodfellow, and K.~Talwar,
  ``Semi-supervised knowledge transfer for deep learning from private training
  data,'' in \emph{ICLR}, 2017.

\bibitem{Neal_18_SemiSupervised_Minimizing_Variance}
N.~Jean, S.~M. Xie, and S.~Ermon, ``Semi-supervised deep kernel learning:
  Regression with unlabeled data by minimizing predictive variance,'' in
  \emph{Advances in Neural Information Processing Systems 31}, S.~Bengio,
  H.~Wallach, H.~Larochelle, K.~Grauman, N.~Cesa-Bianchi, and R.~Garnett,
  Eds.\hskip 1em plus 0.5em minus 0.4em\relax Curran Associates, Inc., 2018,
  pp. 5322--5333.

\bibitem{Abhishek_18_SemiSuper_GANs}
A.~Kumar, P.~Sattigeri, and T.~Fletcher, ``Semi-supervised learning with gans:
  Manifold invariance with improved inference,'' in \emph{Advances in Neural
  Information Processing Systems 30}, I.~Guyon, U.~V. Luxburg, S.~Bengio,
  H.~Wallach, R.~Fergus, S.~Vishwanathan, and R.~Garnett, Eds.\hskip 1em plus
  0.5em minus 0.4em\relax Curran Associates, Inc., 2017, pp. 5534--5544.

\bibitem{Oliver_2018_RealisticSSL}
\BIBentryALTinterwordspacing
A.~Oliver, A.~Odena, C.~A. Raffel, E.~D. Cubuk, and I.~J. Goodfellow,
  ``Realistic evaluation of deep semi-supervised learning algorithms,'' in
  \emph{Advances in Neural Information Processing Systems 31: Annual Conference
  on Neural Information Processing Systems 2018, NeurIPS 2018, 3-8 December
  2018, Montr{\'{e}}al, Canada.}, 2018, pp. 3239--3250. [Online]. Available:
  \url{http://papers.nips.cc/paper/7585-realistic-evaluation-of-deep-semi-supervised-learning-algorithms}
\BIBentrySTDinterwordspacing

\bibitem{Gupta_2018_SemiSupervisSentiment}
R.~Gupta, S.~Sahu, C.~Espy~Wilson, and S.~Narayanan, ``Semi-supervised and
  transfer learning approaches for low resource sentiment classification,'' in
  \emph{Proceedings of ICASSP}, April 2018.

\bibitem{Zhou_2018_SSL_TL}
H.-Y. Zhou, A.~Oliver, J.~Wu, and Y.~Zheng, ``When semi-supervised learning
  meets transfer learning: Training strategies, models and datasets,''
  \emph{arXiv}, vol. abs/1812.05313, 2018.

\bibitem{Tishby99informationBotteleneck}
N.~Tishby, F.~C. Pereira, and W.~Bialek, ``The information bottleneck method,''
  \emph{The 37'th Allerton Conference on Communication, Control, and
  Computing}, 1999.

\bibitem{Shamir_10_InformationBottleneckAnalysis}
O.~Shamir, S.~Sabato, and N.~Tishby, ``Learning and generalization with the
  information bottleneck,'' \emph{Theor. Comput. Sci.}, vol. 411, no. 29-30,
  pp. 2696--2711, 2010.

\bibitem{Friedman_01_Multivariate_IB}
N.~Friedman, O.~Mosenzon, N.~Slonim, and N.~Tishby, ``Multivariate information
  bottleneck,'' in \emph{{UAI} '01: Proceedings of the 17th Conference in
  Uncertainty in Artificial Intelligence, University of Washington, Seattle,
  Washington, USA, August 2-5, 2001}, 2001, pp. 152--161.

\bibitem{Tishby15_information_deeplearning}
\BIBentryALTinterwordspacing
N.~Tishby and N.~Zaslavsky, ``Deep learning and the information bottleneck
  principle,'' in \emph{2015 {IEEE} Information Theory Workshop, {ITW} 2015,
  Jerusalem, Israel, April 26 - May 1, 2015}, 2015, pp. 1--5. [Online].
  Available: \url{https://doi.org/10.1109/ITW.2015.7133169}
\BIBentrySTDinterwordspacing

\bibitem{Shwartz-ZivT17_blackbox}
\BIBentryALTinterwordspacing
R.~Shwartz{-}Ziv and N.~Tishby, ``Opening the black box of deep neural networks
  via information,'' \emph{CoRR}, vol. abs/1703.00810, 2017. [Online].
  Available: \url{http://arxiv.org/abs/1703.00810}
\BIBentrySTDinterwordspacing

\bibitem{Andrew_18_OnTheInformationBottleneck}
A.~M. Saxe, Y.~Bansal, J.~Dapello, M.~Advani, A.~Kolchinsky, B.~D., and
  T.~D.~D. Cox, ``On the information bottleneck theory of deep learning,'' in
  \emph{ICLR}, 2018.

\bibitem{Marylou_18_MI_in_DL}
M.~Gabri\'{e}, A.~Manoel, C.~Luneau, j.~barbier, N.~Macris, F.~Krzakala, and
  L.~Zdeborov\'{a}, ``Entropy and mutual information in models of deep neural
  networks,'' in \emph{Advances in Neural Information Processing Systems 31},
  S.~Bengio, H.~Wallach, H.~Larochelle, K.~Grauman, N.~Cesa-Bianchi, and
  R.~Garnett, Eds.\hskip 1em plus 0.5em minus 0.4em\relax Curran Associates,
  Inc., 2018, pp. 1821--1831.

\bibitem{GeneralizationError_18_Jakubovitz}
\BIBentryALTinterwordspacing
D.~Jakubovitz, R.~Giryes, and M.~R.~D. Rodrigues, ``Generalization error in
  deep learning,'' \emph{arXiv}, vol. abs/1808.01174, 2018. [Online].
  Available: \url{https://arxiv.org/abs/1808.01174}
\BIBentrySTDinterwordspacing

\bibitem{Keskar_2017_Large_Batch_Training}
\BIBentryALTinterwordspacing
N.~S. Keskar, D.~Mudigere, J.~Nocedal, M.~Smelyanskiy, and P.~T.~P. Tang, ``On
  large-batch training for deep learning: Generalization gap and sharp
  minima,'' in \emph{ICLR}, 2017. [Online]. Available:
  \url{https://arxiv.org/pdf/1609.04836.pdf}
\BIBentrySTDinterwordspacing

\bibitem{Kingma15Adam}
\BIBentryALTinterwordspacing
D.~P. Kingma and J.~Ba, ``Adam: {A} method for stochastic optimization,'' in
  \emph{ICLR}, 2015. [Online]. Available: \url{http://arxiv.org/abs/1412.6980}
\BIBentrySTDinterwordspacing

\end{thebibliography}

\end{document}